\documentclass{article}
\usepackage{spconf,amsmath,graphicx}

\usepackage{amssymb}
\usepackage{latexsym}
\usepackage{threeparttable}
\usepackage{booktabs}
\usepackage{algorithm}
\usepackage{algorithmic}
\usepackage{bm}
\usepackage{multirow}
\usepackage{float}
\usepackage{url}
\usepackage{xcolor}
\usepackage{array}
\usepackage{amsmath}
\usepackage{makecell}

\allowdisplaybreaks[4]

\title{Reducing the gap between streaming and non-streaming Transducer-based ASR by adaptive two-stage knowledge distillation}
\name{\begin{tabular}{c} Haitao Tang$^{1}$, Yu Fu$^{2}$, Lei Sun$^{1}$, Jiabin Xue$^{3}$, Dan Liu$^{1}$, Yongchao Li$^{1}$, \\  Zhiqiang Ma$^{1}$, Minghui Wu$^{1}$, Jia Pan$^{1}$, Genshun Wan$^{1}$, Ming'en Zhao$^{1}$\end{tabular}}

\address{$^{1}$ iFLYTEK Research\\
$^{2}$ Zhejiang University\\
$^{3}$ Harbin Institute of Technology\\}
\begin{document}
%
\maketitle
\begin{abstract}
Transducer is one of the mainstream frameworks for streaming speech recognition. There is a performance gap between the streaming and non-streaming transducer models due to limited context. To reduce this gap, an effective way is to ensure that their hidden and output distributions are consistent, which can be achieved by hierarchical knowledge distillation. However, it is difficult to ensure the distribution consistency simultaneously because the learning of the output distribution depends on the hidden one. In this paper, we propose an adaptive two-stage knowledge distillation method consisting of hidden layer learning and output layer learning. In the former stage, we learn hidden representation with full context by applying mean square error loss function. In the latter stage, we design a power transformation based adaptive smoothness method to learn stable output distribution. It achieved 19\% relative reduction in word error rate, and a faster response for the first token compared with the original streaming model in LibriSpeech corpus.
\end{abstract}
\section{Introduction}
\label{sec:intro}
RNN transducer (RNN-T) model~\cite{graves2012sequence, sainath2020streaming} is one of the mainstream frameworks for streaming speech recognition. It usually consists of three modules including encoder, decoder and joint network. Recent studies have popularized using conformer~\cite{gulati2020conformer} which combines convolutional neural network (CNN) with transformer~\cite{zhang2020transformer} as an encoder because it is good at capturing local and global context. However, it brings significantly higher model complexity and latency than transformer due to the macaron structure~\cite{gulati2020conformer}. 


The trade-off between latency and accuracy is a crucial issue in actual scenarios. For example, most of the speech recognition applications require rapid response~\cite{kurata2020knowledge}. Although streaming model can respond much faster than the non-streaming one, it usually performs significantly worse due to limited context~\cite{kojima2021knowledge}. Hierarchical knowledge distillation (KD) is an effective strategy to solve this issue, which reduces the differences between the hidden and output distributions of streaming and non-streaming models~\cite{kojima2021knowledge}. However, it is difficult for streaming model to learn these distributions from the non-streaming simultaneously because learning of output distribution depends on the hidden one.
\begin{figure*}[t]
 \centering
 \centerline{\includegraphics[width=14cm]{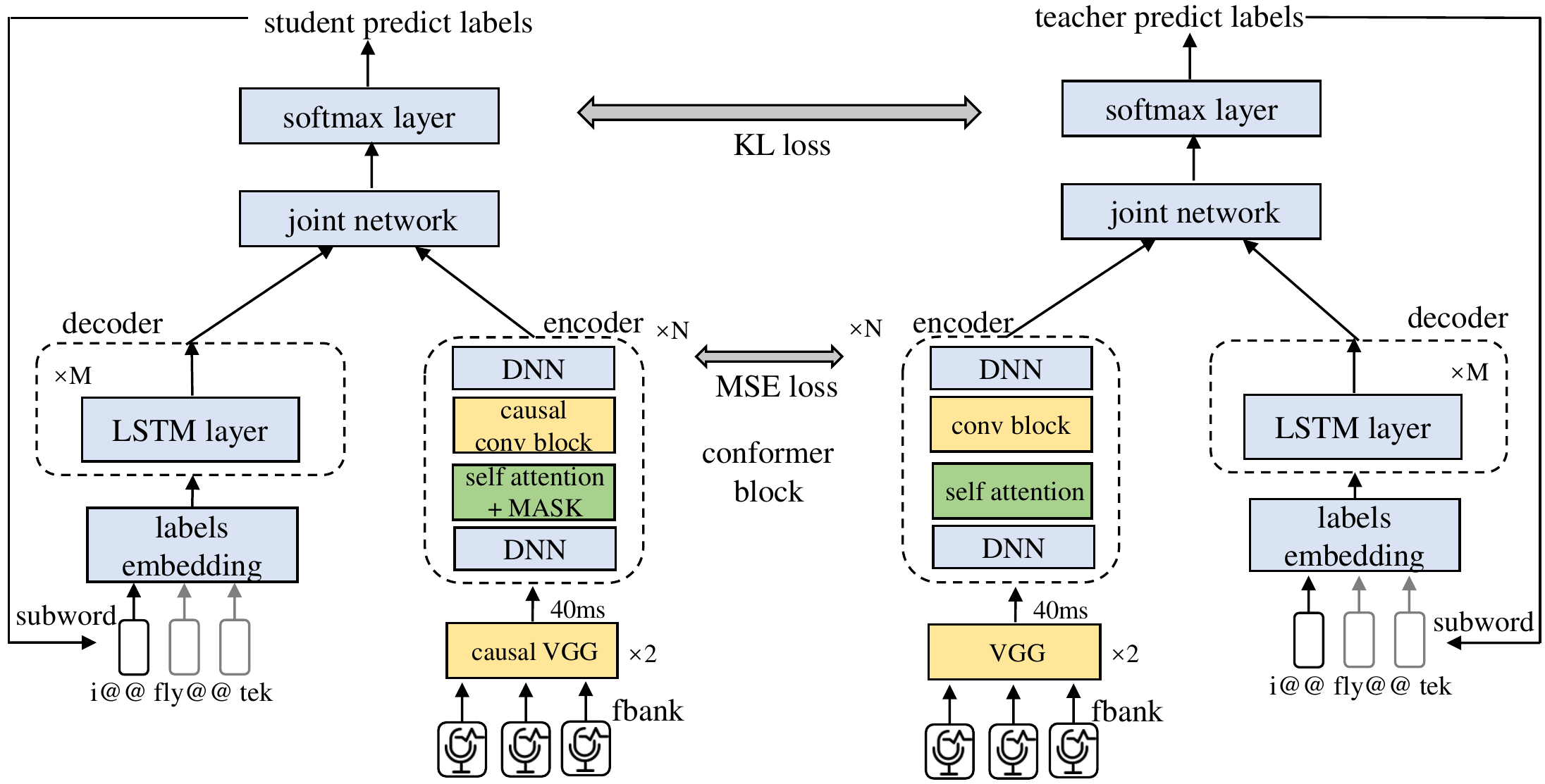}}
\caption{The whole framework of the adaptive two-stage knowledge distillation method for conformer transducer.} 
\label{ICASSP_framework}
\end{figure*}

In this paper, we propose a novel adaptive hierarchical knowledge distillation method with a two-stage architecture. It ensures learning of output distribution depends on the hidden one. In the first stage, because hidden distribution is more necessary to learn comparing with the output one, we learn a hidden representation with full context by applying the mean square error (MSE) loss function~\cite{jiao2019tinybert}. After this stage, it can focus more on learning output distribution of the non-streaming model. The Kullback-Leibler (KL) divergence~\cite{romero2014fitnets} is employed to minimize their output distribution distances in the latter stage. Furthermore, streaming model needs to learn a smoothed soft target distribution for each case by introducing the temperature coefficient in KL divergence~\cite{hinton2015distilling}. Because the output distribution of each case has different sharpness, it is difficult to obtain a smooth distribution by adjusting the above coefficient manually~\cite{rao2022parameter}. To address this issue, we design an adaptive temperature method based power transformation to control the smoothness of the output distribution dynamically. After that, streaming model can learn a stable output distribution from the non-streaming more efficiently.

\section{System description}
\label{sec:format}

\subsection{Hierarchical knowledge distillation based transducer}

The conformer transducer (CT) is an end-to-end speech recognition model. Let acoustic feature $\mathbf{X}$ = $[\mathbf{x}_{1},\mathbf{x}_{2},\cdots,\mathbf{x}_{T}]$, previous token $\mathbf{Y}$ = $[{y}_{1},{y}_{2},\cdots,{y}_{U}]$, $\mathbf{h}_{t}^{enc}$ is the acoustic representation extracted by the acoustic encoder at time $t$, and $\mathbf{h}_{u}^{dec}$ is text representation of the $u$-th subword unit by the decoder network. Then, the output of joint network, $\mathbf{h}_{t,u}^{joint}$,  is fed to the output layer to compute the probability distribution. We present detailed steps for the CT in Fig.~\ref{ICASSP_framework}, which can be written as follows:
\begin{small}
\begin{align}
\mathbf{h}_{t}^{enc} &= encoder(\mathbf{x}_{t}) \\
\mathbf{h}_{u}^{dec} &= decoder({y}_{u}) \\
\mathbf{h}_{t,u}^{joint} &= joint(\mathbf{h}_{t}^{enc}, \mathbf{h}_{u}^{dec}) \\
{P}(\hat{y}_{t+1}|{x}_{t},{y}_{u}) &= softmax(\mathbf{h}_{t,u}^{joint})
\end{align}
\end{small}


\noindent where ${encoder(\cdot)}$ is composed of VGG~\cite{yeh2019transformer} and conformer~\cite{ zeineldeen2022conformer}. ${decoder(\cdot)}$ is a multi-layer LSTM~\cite{rao2017exploring}, and joint network, ${joint(\cdot)}$, consists of fully connected (FC) layer~\cite{graves2013speech}. Furthermore, the streaming model employs causal convolution in VGG and conv block of conformer layers, truncated left 16 and right 0 frames by simple masks in self-attention~\cite{chen2021developing}. We should note that streaming model performs significantly worse than the non-streaming model due to limited context~\cite{kojima2021knowledge}, so hierarchical KD is adopted to solve this issue. It employs MSE loss function in encoder-decoder layer outputs and KL divergence in joint network outputs (grey double arrow in Fig.~\ref{ICASSP_framework}). In this manner, the streaming model distills the required knowledge from the non-streaming model. 

\subsection{Adaptive two-stage KD}

The streaming model in hierarchical KD is hard to learn hidden and output distributions simultaneously from non-streaming model, because the latter distributions depend on the former one. To solve this problem, we propose an adaptive hierarchical KD method with two-stage. Here, we define two tasks, hidden layer learning and output layer learning, to better describe the two-stage KD. Streaming model is student and non-streaming model is teacher.

~\textbf{Hidden layer learning.} To learn intermediate representations of non-streaming network, encoder-decoder of the non-streaming model applies the MSE loss function (grey double arrow between the dashed-line boxes in Fig.~\ref{ICASSP_framework}) to guide all encoder and decoder layers of the streaming model:
\begin{small}
\begin{equation}
\label{hidden}
\mathcal{L}_{hidden} = \sum_{i=1}^{N}{MSE}(\mathbf{E}_{i}^{S}, \mathbf{E}_{i}^{T}) + \sum_{j=1}^{M}{MSE}(\mathbf{D}_{j}^{S}, \mathbf{D}_{j}^{T})
\end{equation}
\end{small}

\noindent where ${MSE(\cdot)}$ is the MSE loss function~\cite{jiao2019tinybert}, $\mathbf{E}_{i}^{S} \in \mathbb{R}^{T_i \times d_i}$ and $\mathbf{E}_{i}^{T} \in \mathbb{R}^{T_i \times d_i}$ are outputs of the ${i}$-th conformer layer for streaming and non-streaming. $\mathbf{D}_{j}^{S} \in \mathbb{R}^{U_j \times d_j}$ and $\mathbf{D}_{j}^{T} \in \mathbb{R}^{U_j \times d_j}$ are outputs of the ${j}$-th LSTM layer for streaming and non-streaming. The scalar values $d_i$ and $d_j$ denote hidden sizes of conformer and LSTM. $N$ and $M$ represent the number of layers. Usually, hidden representations of the lower layer contain the information on speaker characteristics, and the higher layer contains linguistic content information~\cite{li2020does}. In this way, hidden layers of the streaming model can be distilled from different types of hidden representations of the non-streaming one.


~\textbf{Output layer learning.} In addition to imitating the behaviors of intermediate layers, we also utilize traditional knowledge distillation to fit output of the non-streaming model. The streaming model is trained to optimize:
\begin{small}
\begin{equation}
\label{loss_function}
\mathcal{L}_{output} = \mathcal{L}_{rnnt}(\mathbf{Q}^{S}, {y}_{true}) + {KL}(\mathbf{Q}^{S} / \tau, \mathbf{Q}^{T} / \tau)
\end{equation}
\end{small}

\noindent where $\mathbf{Q}^{S} \in \mathbb{R}^{T \times U \times V}$ and $\mathbf{Q}^{T} \in \mathbb{R}^{T \times U \times V}$ are output of Eq. (4) of streaming and non-streaming models. $V$ is dictionary size. Note that the first term in Eq.~(\ref{loss_function}) corresponds to transducer loss of the streaming model. The second term enforces the streaming model to learn soft target distribution from the non-streaming one. $\tau$ is the temperature. We set $\tau = 1$.

However, streaming model needs to learn a smoothed soft target distribution for each case by introducing the temperature coefficient in KL divergence~\cite{hinton2015distilling}. 
Because the output distribution of each case has different sharpness, it is difficult to obtain a smooth distribution by adjusting the above coefficient manually~\cite{rao2022parameter}. 
To control the smoothness of the output distribution dynamically, we attempt to obtain a lossless smooth distribution, which has two key characteristics, i.e. same informativeness and consistent monotonicity, comparing original distribution. Considering entropy can not only ensure monotonicity of two distributions but also measure information loss, we employ it to constrain the difference of before and after smoothing. For this purpose, we define a novel smooth distribution $\widetilde{\mathbf{Q}}$ as:


\begin{small}
\begin{equation}
\label{entropy}
\widetilde{\mathbf{Q}} = -\sum_{v=1}^{V}{\mathbf{Q}_{v} \log \mathbf{Q}_{v}} = \mathbb{E}(-\log \mathbf{Q}_{v})
\end{equation}
\end{small}

\noindent where $\mathbf{Q} \in \mathbb{R}^{T \times U \times V}$ is output of softmax layer after joint network, ${Q}_{t, u, v} \in [0, 1]$, $\widetilde{{Q}}_{t, u, v} \in [0, \log V]$. 

Furthermore, to ensure that entropy output is in the $[0, \log V]$ range, we attempt to smooth the output distribution by applying a power transformation on ${Q}_{t, u, v}$:

\begin{small}
\begin{equation}
\label{entropy_power}
\begin{aligned}
\widetilde{\mathbf{Q}}^{\gamma} &= -\sum_{v=1}^{V}{\frac{\mathbf{Q}^{\gamma}_{v}}{\sum_{v=1}^{V}{\mathbf{Q}^{\gamma}_{v}}} \log \frac{\mathbf{Q}^{\gamma}_{v}}{\sum_{v=1}^{V}{\mathbf{Q}^{\gamma}_{v}}}}\\
&= \log \sum_{v=1}^{V}{\mathbf{Q}^{\gamma}_{v}} - \frac{\gamma \sum_{v=1}^{V}\mathbf{Q}^{\gamma}_{v} \log \mathbf{Q}_{v}}{\sum_{v=1}^{V}{\mathbf{Q}^{\gamma}_{v}}}
\end{aligned}
\end{equation}
\end{small}

\noindent where $\gamma$ is adaptive temperature, $\widetilde{{Q}}^{\gamma}_{t, u, v}$ is the maximum value $\log V$, if $\gamma = 0$, and $\widetilde{{Q}}^{\gamma}_{t, u, v}$ is the minimum value $0$, if $\gamma \xrightarrow[]{} \infty$. Thus, we need to compute $\gamma$ in the $[0, \log V]$ transformation range. To make solution easier, $\mathbf{Q}^{\gamma}_{v} \approx \mathbf{Q}_{v} + (\gamma - 1) \mathbf{Q}_{v} \log \mathbf{Q}_{v}$. $\widetilde{\mathbf{Q}}^{\gamma}$ is approximated by Taylor series of $\gamma = 1$:
\begin{small}
\begin{equation}
\label{entropy_power_taylor}
\begin{aligned}
\widetilde{\mathbf{Q}}^{\gamma} &\approx -\sum_{v=1}^{V}{\mathbf{Q}_{v} \log \mathbf{Q}_{v}} \\&+((\sum_{v=1}^{V} {\mathbf{Q}_{v} \log \mathbf{Q}_{v}})^{2} -\sum_{v=1}^{V}{\mathbf{Q}_{v} \log (\mathbf{Q}_{v})^2})(\gamma - 1)\\
&= \widetilde{\mathbf{Q}} + (\widetilde{\mathbf{Q}}^{2} - \mathbb{E}[(\log \mathbf{Q}_{v})^{2}])(\gamma - 1)
\end{aligned}
\end{equation}
\end{small}

Therefore, $\gamma$ is computed as:

\begin{small}
\begin{equation}
\label{gamma}
\gamma \approx 1 + \frac{\widetilde{\mathbf{Q}}^{\gamma} - \widetilde{\mathbf{Q}}}{\widetilde{\mathbf{Q}}^{2} - \mathbb{E}[(\log \mathbf{Q}_{v})^{2}]}
\end{equation}
\end{small}

\noindent where $\widetilde{\mathbf{Q}}^{\gamma}$ is the expected entropy, which is fixed as the maximum value, $\log V$, to ensure the molecule in Eq.~(\ref{gamma}) $ \ge 0$. Because ${Q}_{t, u, v} \in [0, 1]$, the first term in the denominator is smaller than the second one, $\gamma \in [0, 1]$. When $\gamma$ is lower, the $\mathbf{Q}^{\gamma}_{v}$ is more smoother in streaming and non-streaming models.

In this manner, we can use iterative solution in maximum step size $Z$ to obtain optimal $\gamma$ and $\mathbf{Q}^{Z}$. The detailed steps are described in Algorithm \ref{algorithm:2}. Thus Eq.~(\ref{loss_function}) can be converted to:
\begin{small}
\begin{equation}
\label{output_learning}
\mathcal{L}_{output} = \mathcal{L}_{rnnt}(\mathbf{Q}^{S}, {y}_{true}) + {KL}({\mathbf{Q}}^{S,Z}, {\mathbf{Q}}^{T,Z})
\end{equation}
\end{small}

\noindent where ${\mathbf{Q}}^{S,Z}$ and ${\mathbf{Q}}^{T,Z}$ are the power transformation output of streaming and non-streaming.

~\textbf{Two-stage KD.} In order to ensure that output learning depends on the hidden one, we unify distillation loss of corresponding layers by combining the above two tasks, i.e., Eq.~(\ref{hidden}) and Eq.~(\ref{output_learning}). Therefore, two-stage KD loss function can be written as:
\begin{small}
\begin{equation}
\label{kd}
\mathcal{L}_{KD} = \alpha \times \mathcal{L}_{hidden} + \beta \times \mathcal{L}_{output}
\end{equation}
\end{small}

\noindent where $\alpha$ and $\beta$ are hyper-parameters to balance two tasks. Because it is challenging for the tasks to achieve their aims simultaneously, we design an adaptive two-stage learning framework based on both main-task and sub-task. When one hyper-parameter is much larger (e.g., $1\times10^{1}$, $1\times10^{2}$, $1\times10^{3}$) than another, its corresponding task will switch as main-task to guide the training of streaming model, and the other one will switch as a sub-task to assist the training.

\renewcommand{\algorithmicrequire}{\textbf{Input:}}
\renewcommand{\algorithmicensure}{\textbf{Output:}}
\renewcommand{\thealgorithm}{1}
\begin{algorithm}[!t]
\caption{Computation of power transformation.}
\label{algorithm:2}
\begin{algorithmic}[1]
\REQUIRE{$\widetilde{\mathbf{Q}}^{\gamma}$, $\mathbf{Q}$, $Z$}
\STATE Initialization $\mathbf{Q}^0 = \mathbf{Q}$.
\FOR{$ z = 1,2,...,Z$}
\STATE $\widetilde{\mathbf{Q}} = -\sum_{v=1}^{V}{\mathbf{Q}_{v}^{z-1} \log \mathbf{Q}_{v}^{z-1}}$
\STATE $\gamma = 1 + \frac{(\widetilde{\mathbf{Q}}^{\gamma} - \widetilde{\mathbf{Q}})}{\widetilde{\mathbf{Q}}^2-\sum_{v=1}^{V}{(\mathbf{Q}_{v}^{z - 1} \times (log\mathbf{Q}_{v}^{z - 1})^2)}}$
\STATE $\mathbf{Q}^{z} = \frac{{(\mathbf{Q}^{z-1})}^{\gamma}}{\sum_{v=1}^{V}{{(\mathbf{Q}_{v}^{z-1})}^{\gamma}}}$
\ENDFOR
\ENSURE{$\mathbf{Q}^{z}$}
\end{algorithmic}
\end{algorithm}

Considering hidden layer is more necessary to learn comparing output layer. Thus we apply the larger $\alpha$ and smaller $\beta$ in the first stage so that the hidden layer learning is main-task and the output layer learning is sub-task. However, results of the first stage perform generally worse than the original streaming due to lacking alignment capability of acoustic features and output tokens.


To solve the above problem, output layer learning is switched to main-task and hidden layer learning switches to sub-task in the second stage. In this manner, the streaming model mainly learns output distribution of the non-streaming model, and keeps the capacity that is predictive of intermediate representations of the non-streaming model.
\section{experiments and result analysis}
\label{sec:experiments}

\subsection{Experimental setups}
Regarding experimental dataset, we used the LibriSpeech corpus~\cite{panayotov2015librispeech}. We applied log Mel filter bank as input acoustic features. Word error rate (WER) and relative WER reduction (WERR) are employed as evaluation metrics. The encoder contains 12-layers conformer as described in \cite{zhang2022ustc}. The decoder consists of an embedding and an LSTM layer. For both streaming and non-streaming models, we applied same architecture of 73M parameters. In the first stage, $\alpha = 1$, $\beta = 0.01$, and in the second stage, $\alpha = 0.01$, $\beta = 1$ .
\subsection{Results}
Table~\ref{compare_SOTA} summarizes the proposed and state-of-the-art (SOTA) methods~\cite{kojima2021knowledge} (denoted by S1, S2, S6). The original streaming model was trained without KD. The traditional KD corresponds to Eq.~(\ref{loss_function}). The hierarchical KD corresponds to Eq.~(\ref{kd}), where $\alpha = \beta = 1$. Compared with original streaming model, results of traditional KD and hierarchical KD have slightly higher recognition accuracy in the test set. In the proposed method, results of the first stage generally perform worse than original streaming due to lacking alignment capability between acoustic features and output tokens. However, combined with the first and second stages, the streaming model improved 17.05\% in WERR comparing original. By learning output distributions of different smoothness, the performance of adaptive two-stage KD is further increased with 2.31\% and 2.19\% in WERR of test-clean and test-other, respectively. The SOTA methods in Table~\ref{compare_SOTA} include streaming (S2), non-streaming (S1), and streaming with hierarchical KD (S6)~\cite{kojima2021knowledge}. Relative to the S6 which achieves 2.10\% in WERR, the proposed method can substantially reduce gap between streaming and non-streaming models with 19.24\% in WERR. It is worth mentioning that the non-streaming method achieved similar WER to the other SOTA~\cite{panchapagesan2021efficient}.


\begin{table}
\centering
\caption{Results of adaptive two-stage KD and SOTA.} 
 \setlength{\tabcolsep}{1.0 mm}{
\begin{tabular}{lcccc}
\toprule
\multirow{2}{*}{Description} & \multicolumn{2}{c}{test-clean} & \multicolumn{2}{c}{test-other} \\ \cline{2-5} 
                    & WER                & WERR      & WER                & WERR  \\ \hline
original streaming  & 3.90               & 0.00      & 8.21               & 0.00      \\
non-streaming       & 2.42               & 37.94     & 4.35               & 47.01     \\
traditional KD      & 3.49               & 10.51     & 7.92               & 3.53      \\
hierarchical KD     & 3.39               & 13.07     & 7.46               & 9.13      \\
two-stage KD(first)      & 5.46               & -40.00    & 12.16              & -48.11    \\
two-stage KD(second)     & 3.27               & 16.15     & 6.81               & 17.05     \\ 
two-stage KD(adaptive)   & \textbf{3.18}     & \textbf{18.46} & \textbf{6.63}      & \textbf{19.24}     \\ \hline
original streaming S2   & -                  & 0.00           & -                  & 0.00      \\
non-streaming S1        & -                  & 25.40          & -                  & 31.80     \\
hidden KD S6            & -                  & 3.50           & -                  & 2.10      \\ 
\bottomrule
\end{tabular}}
\label{compare_SOTA}
\end{table}

\begin{table}
\centering
\caption{Effectiveness of sub-task coefficient.}
 \setlength{\tabcolsep}{1.0 mm}{
\begin{tabular}{lcc}
\toprule
WER(\%)                     & test-clean & test-other \\ \hline
original streaming          & 3.90       & 8.21       \\
non-streaming               & 2.42       & 4.35       \\
two-stage KD(0.1)           & 3.16       & 6.89       \\
two-stage KD(0.01)          & \textbf{3.18}       & \textbf{6.63}       \\
two-stage KD(0.001)         & 3.17       & 7.22       \\
two-stage KD-fixed(0.01)      & 4.86       & 10.8       \\
\bottomrule
\end{tabular}}
\label{tune_param}
\end{table}

We compared effectiveness of sub-task coefficient in Table~\ref{tune_param}. In adaptive two-stage KD, hyper-parameter of the main-task is always set as 1, while hyper-parameter in the sub-task is set as a small constant (e.g., we have tried 0.1, 0.01 and 0.001). Based on that, we find that $\alpha=1, \beta=0.01$ in the first stage and $\alpha=0.01, \beta=1$ in second stage are the optimal configuration of hyper-parameters. In the second stage, streaming model needs to keep capacity that is predictive of intermediate representations of non-streaming network. In addition to fine-tuning the sub-task by a small constant, we also fixed encoder and decoder parameters trained in the first stage to maintain this capacity better. The fixed parameters method in the second stage (see last line in Table~\ref{tune_param}) is only fine-tunes joint network by Eq.~(\ref{output_learning}), which performs worse than original streaming.




To prove the universality of our proposed method, we designed three different CT models with parameters of 73M, 36M and 18M by modifying conformer layers and hidden units. We analyzed different parameters in WER and response time about the first token in Table~\ref{structure_latency}.  The response time about the first token is measured on the test-clean. With model parameters increased, the WER decreased. Compared with original streaming model, the WERR of distilled streaming CT is reduced by 19.24\%, 18.10\% and 15.97\% respectively, on the test-other set. In same parameter size, the response time about the first token in distilled streaming model could be faster than original streaming model because it had learned full hierarchical context from non-streaming one.

\begin{table}
\centering
\caption{Results of different parameters and latency.}
 \setlength{\tabcolsep}{1.0 mm}{
\begin{tabular}{lcccc}
\toprule
\multicolumn{2}{c}{\multirow{2}{*}{param}} & \multicolumn{2}{c}{WER(\%)} & \multirow{2}{*}{\begin{tabular}[c]{@{}c@{}}response time\\ (ms)\end{tabular}} \\ \cline{3-4}
\multicolumn{2}{c}{}                       & test-clean& test-other      &                                                                         \\ \hline
\multirow{3}{*}{73M}     & original streaming   & 3.90      & 8.21            & 933                                                                     \\
                         & non-streaming   & 2.42      & 4.35            & -                                                                     \\
                         & two-stage KD    & 3.18      & 6.63            & 905                                                                     \\ \hline
\multirow{3}{*}{36M}     & original streaming   & 4.98      & 11.49           & 908                                                                     \\
                         & non-streaming   & 3.01      & 6.30            & -                                                                     \\
                         & two-stage KD    & 4.17      & 9.41            & 875                                                                     \\ \hline
\multirow{3}{*}{18M}     & original streaming   & 7.31      & 16.34           & 890                                                                     \\
                         & non-streaming   & 4.37      & 9.65            & -                                                                     \\
                         & two-stage KD    & 5.83      & 13.73           & 865                                                                     \\ 
\bottomrule
\end{tabular}}
\label{structure_latency}
\end{table}

\vspace{-0.15cm}
\section{Conclusions}
\vspace{-0.15cm}
\label{sec:conclusion}
In this work, we focused on reducing gap between the streaming and non-streaming models. To solve this issue, we proposed an adaptive two-stage knowledge distillation method. It ensures that learning of the output distribution depends on that of the hidden one. In the first stage, we learned a hidden representation with full context by applying the MSE loss function. Then the KL divergence was employed to minimize their output distribution distances in the latter stage. We also designed an adaptive temperature method based power transformation to dynamically control the smoothness of the output distribution. We compared the proposed method with SOTA methods and observed 19.24\% relative reduction in WER and a faster response for the first token.



\vfill\pagebreak



\bibliographystyle{IEEEbib}
\bibliography{strings,refs}

\begin{thebibliography}{10}

\bibitem{graves2012sequence}
Alex Graves,
\newblock ``Sequence transduction with recurrent neural networks,''
\newblock {\em CoRR}, vol. abs/1211.3711, 2012.

\bibitem{sainath2020streaming}
Tara~N. Sainath, Yanzhang He, Bo~Li, et~al.,
\newblock ``A streaming on-device end-to-end model surpassing server-side
  conventional model quality and latency,''
\newblock in {\em 2020 {IEEE} International Conference on Acoustics, Speech and
  Signal Processing, {ICASSP} 2020, Barcelona, Spain, May 4-8, 2020}. 2020, pp.
  6059--6063, {IEEE}.

\bibitem{gulati2020conformer}
Anmol Gulati, James Qin, Chung{-}Cheng Chiu, et~al.,
\newblock ``Conformer: Convolution-augmented transformer for speech
  recognition,''
\newblock pp. 5036--5040, 2020.

\bibitem{zhang2020transformer}
Qian Zhang, Han Lu, Hasim Sak, et~al.,
\newblock ``Transformer transducer: {A} streamable speech recognition model
  with transformer encoders and {RNN-T} loss,''
\newblock in {\em 2020 {IEEE} International Conference on Acoustics, Speech and
  Signal Processing, {ICASSP} 2020, Barcelona, Spain, May 4-8, 2020}. 2020, pp.
  7829--7833, {IEEE}.

\bibitem{kurata2020knowledge}
Gakuto Kurata and George Saon,
\newblock ``Knowledge distillation from offline to streaming {RNN} transducer
  for end-to-end speech recognition,''
\newblock in {\em Interspeech 2020, 21st Annual Conference of the International
  Speech Communication Association, Virtual Event, Shanghai, China, 25-29
  October 2020}, Helen Meng, Bo~Xu, and Thomas~Fang Zheng, Eds. 2020, pp.
  2117--2121, {ISCA}.

\bibitem{kojima2021knowledge}
Atsushi Kojima,
\newblock ``Knowledge distillation for streaming transformer-transducer,''
\newblock in {\em Interspeech 2021, 22nd Annual Conference of the International
  Speech Communication Association, Brno, Czechia, 30 August - 3 September
  2021}, Hynek Hermansky, Honza Cernock{\'{y}}, Luk{\'{a}}s Burget, Lori Lamel,
  Odette Scharenborg, and Petr Motl{\'{\i}}cek, Eds. 2021, pp. 2841--2845,
  {ISCA}.

\bibitem{jiao2019tinybert}
Xiaoqi Jiao, Yichun Yin, Lifeng Shang, et~al.,
\newblock ``Tinybert: Distilling {BERT} for natural language understanding,''
\newblock vol. {EMNLP} 2020, pp. 4163--4174, 2020.

\bibitem{romero2014fitnets}
Adriana Romero, Nicolas Ballas, Samira~Ebrahimi Kahou, et~al.,
\newblock ``Fitnets: Hints for thin deep nets,''
\newblock 2015.

\bibitem{hinton2015distilling}
Geoffrey~E. Hinton, Oriol Vinyals, and Jeffrey Dean,
\newblock ``Distilling the knowledge in a neural network,''
\newblock {\em CoRR}, vol. abs/1503.02531, 2015.

\bibitem{rao2022parameter}
Jun Rao, Xv~Meng, Liang Ding, et~al.,
\newblock ``Parameter-efficient and student-friendly knowledge distillation,''
\newblock {\em CoRR}, vol. abs/2205.15308, 2022.

\bibitem{yeh2019transformer}
Ching{-}Feng Yeh, Jay Mahadeokar, Kaustubh Kalgaonkar, et~al.,
\newblock ``Transformer-transducer: End-to-end speech recognition with
  self-attention,''
\newblock {\em CoRR}, vol. abs/1910.12977, 2019.

\bibitem{zeineldeen2022conformer}
Mohammad Zeineldeen, Jingjing Xu, Christoph L{\"{u}}scher, et~al.,
\newblock ``Conformer-based hybrid {ASR} system for switchboard dataset,''
\newblock in {\em {IEEE} International Conference on Acoustics, Speech and
  Signal Processing, {ICASSP} 2022, Virtual and Singapore, 23-27 May 2022}.
  2022, pp. 7437--7441, {IEEE}.

\bibitem{rao2017exploring}
Kanishka Rao, Hasim Sak, and Rohit Prabhavalkar,
\newblock ``Exploring architectures, data and units for streaming end-to-end
  speech recognition with rnn-transducer,''
\newblock 2018, vol. abs/1801.00841.

\bibitem{graves2013speech}
Alex Graves, Abdel{-}rahman Mohamed, and Geoffrey~E. Hinton,
\newblock ``Speech recognition with deep recurrent neural networks,''
\newblock in {\em {IEEE} International Conference on Acoustics, Speech and
  Signal Processing, {ICASSP} 2013, Vancouver, BC, Canada, May 26-31, 2013}.
  2013, pp. 6645--6649, {IEEE}.

\bibitem{chen2021developing}
Xie Chen, Yu~Wu, Zhenghao Wang, et~al.,
\newblock ``Developing real-time streaming transformer transducer for speech
  recognition on large-scale dataset,''
\newblock in {\em {IEEE} International Conference on Acoustics, Speech and
  Signal Processing, {ICASSP} 2021, Toronto, ON, Canada, June 6-11, 2021}.
  2021, pp. 5904--5908, {IEEE}.

\bibitem{li2020does}
Chung{-}Yi Li, Pei{-}Chieh Yuan, and Hung{-}yi Lee,
\newblock ``What does a network layer hear? analyzing hidden representations of
  end-to-end {ASR} through speech synthesis,''
\newblock in {\em 2020 {IEEE} International Conference on Acoustics, Speech and
  Signal Processing, {ICASSP} 2020, Barcelona, Spain, May 4-8, 2020}. 2020, pp.
  6434--6438, {IEEE}.

\bibitem{panayotov2015librispeech}
Vassil Panayotov, Guoguo Chen, Daniel Povey, et~al.,
\newblock ``Librispeech: An {ASR} corpus based on public domain audio books,''
\newblock in {\em 2015 {IEEE} International Conference on Acoustics, Speech and
  Signal Processing, {ICASSP} 2015, South Brisbane, Queensland, Australia,
  April 19-24, 2015}. 2015, pp. 5206--5210, {IEEE}.

\bibitem{zhang2022ustc}
Weitai Zhang, Zhongyi Ye, Haitao Tang, et~al.,
\newblock ``The {USTC-NELSLIP} offline speech translation systems for {IWSLT}
  2022,''
\newblock in {\em Proceedings of the 19th International Conference on Spoken
  Language Translation, IWSLT@ACL 2022, Dublin, Ireland (in-person and online),
  May 26-27, 2022}, Elizabeth Salesky, Marcello Federico, and Marta
  Costa{-}juss{\`{a}}, Eds. 2022, pp. 198--207, Association for Computational
  Linguistics.

\bibitem{panchapagesan2021efficient}
Sankaran Panchapagesan, Daniel~S. Park, Chung{-}Cheng Chiu, et~al.,
\newblock ``Efficient knowledge distillation for rnn-transducer models,''
\newblock in {\em {IEEE} International Conference on Acoustics, Speech and
  Signal Processing, {ICASSP} 2021, Toronto, ON, Canada, June 6-11, 2021}.
  2021, pp. 5639--5643, {IEEE}.

\end{thebibliography}

\end{document}